\documentclass[a4paper,twoside]{article}

\usepackage{graphicx}
\usepackage{float}
\usepackage{multirow}
\usepackage{amsmath,amssymb} % define this before the line numbering.
\usepackage{hyperref}
\usepackage{graphicx}
\usepackage{amsmath}
\usepackage{subcaption}
\usepackage{color,soul}
\usepackage{multirow}
\usepackage{rotating}

\usepackage[width=122mm,left=12mm,paperwidth=146mm,height=193mm,top=12mm,paperheight=217mm]{geometry}

\usepackage{epsfig}
\usepackage{calc}
\usepackage{amssymb}
\usepackage{amstext}
\usepackage{amsmath}
\usepackage{amsthm}
\usepackage{multicol}
\usepackage{pslatex}
\usepackage{apalike}
\usepackage{SCITEPRESS}     % Please add other packages that you may need BEFORE the SCITEPRESS.sty package.

\usepackage{booktabs} 

\begin{document}

\title{Design of Real-time Semantic Segmentation Decoder \\ for Automated Driving}

\author{\authorname{Arindam Das\sup{1}, Saranya Kandan\sup{1}, Senthil Yogamani\sup{2}, Pavel K\v{r}\'{i}\v{z}ek\sup{3}}
 \affiliation{\sup{1}Detection Vision Systems, Valeo India}
 \affiliation{\sup{2}Valeo Vision Systems, Valeo Ireland}
 \affiliation{\sup{3}Valeo R\&D DVS, Prague, Czech Republic }
 \email{\{arindam.das, saranya.kandan, senthil.yogamani, pavel.krizek\}@valeo.com}
}


\abstract{Semantic segmentation remains a computationally intensive algorithm for embedded deployment even with the rapid growth of computation power. Thus efficient network design is a critical aspect especially for applications like automated driving which requires real-time performance.  Recently, there has been a lot of research on designing efficient encoders that are mostly task agnostic.  Unlike image classification and bounding box object detection tasks, decoders are computationally expensive as well for semantic segmentation task. In this work, we focus on efficient design of the segmentation decoder and assume that an efficient encoder is already designed to provide shared features for a multi-task learning system. We design a novel efficient non-bottleneck layer and a family of decoders which fit into a small run-time budget using VGG10 as efficient encoder. We demonstrate in our dataset that experimentation with various design choices led to an improvement of 10\% from a baseline performance.}

\onecolumn \maketitle \normalsize \vfill

\section{Introduction}

Semantic segmentation provides complete semantic scene understanding wherein each pixel in an image is assigned a class label. It has applications in various fields including automated driving, augmented reality and medical image processing.  The complexity of Convolution Neural Networks (CNN) architectures have been growing consistently. However for industrial applications, there is a computational bound because of limited resources on embedded platforms. It is essential to design efficient models which fit the run-time budget of the system. There are many papers which demonstrate large run-time improvements with minimal loss of accuracy by using various techniques. An overview of efficient CNN for semantic segmentation is provided in \cite{briot2018analysis}.

%\section{Background} \label{background}

\begin{figure*}
\centering
\includegraphics[scale= 0.50]{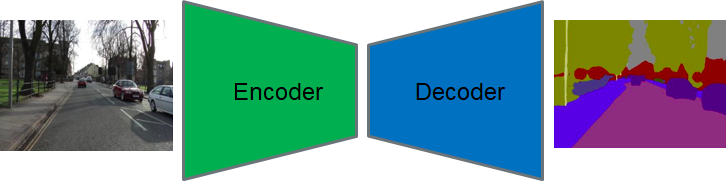}
\caption{Typical encoder-decoder architecture of CNN based semantic segmentation network.}
\label{fig:enc-dec}
\end{figure*}

Semantic segmentation architectures typically have an encoder and a decoder as shown in Figure \ref{fig:enc-dec}. The encoder extracts features from the image which is then decoded to produce semantic segmentation output. ImageNet pre-trained networks are typically used as encoder. In early architectures  \cite{badrinarayanan2015segnet} \cite{ronneberger2015u}, decoder was a mirror image of encoder and had the same complexity. Newer architectures use a relatively smaller decoder. There can also be additional connections from encoder to decoder. For example, SegNet \cite{badrinarayanan2015segnet} passes max-pooling indices and Unet \cite{ronneberger2015u} passes intermediate feature maps to decoder.

In this paper, we propose the design of a novel non-bottleneck layer particularly to perform semantic segmentation task where the encoder is task independent unlike existing methods. Our non-bottleneck layer based on residual learning, has been designed to perform well for some classes that are not well represented in the dataset. Having cascaded skip connections make our non-bottleneck layer capable to handle high gradient flow and suitable for an embedded platform to run on real time with lightweight encoder model. Table \ref{table1} presents the efficacy of the proposed decoder architecture over many variants of the decoder model while maintaining the encoder to be task independent and constant.

The rest of the paper is structured as follows. 
%Section \ref{background} provides the background and related work on semantic segmentation. 
Section \ref{background} discusses the related work on semantic segmentation. \ref{proposed} explains the details of the proposed network. Section \ref{results} details the experimental setup and results. Finally, section \ref{conclusion} summarizes the paper and provides potential future directions.

\section{Background} \label{background}

The main purpose of convolutional neural network was to perform image classification so that an object of a particular category can be predicted irrespective of its rotation, translation and scale. But Long et \textit{al.} \cite{long2015fully} considered the same concept as encoder and added fully convolutional upsampling layer using the concept of unpooling to construct the output. Later deconvolution technique was introduced instead of unpooling in \cite{badrinarayanan2015segnet} along with considering skip connections to pull encoded feature maps. As part of the recent advancement in semantic segmentation tasks, high usage of dilated convolution is observed in many works \cite{chen2018deeplab} \cite{chen2018encoder} \cite{DBLP:journals/corr/abs-1709-00179} \cite{romera2017efficient} \cite{wang2018understanding} \cite{yu2015multi} \cite{zhao2017icnet} \cite{yang2018denseaspp}, this is because it helps to grow the receptive field exponentially without loss of resolution. As notable other works, DeepLab2 \cite{chen2018deeplab} uses spatial pyramid pooling in ResNet-101 and it includes Conditional Random Fields (CRF) as well. RefineNet \cite{lin2017refinenet} proposed a multi-path refinement network that exploits information to perform better segmentation in higher resolution. It basically refines features from low resolution in recursive manner to obtain better prediction in high resolution. Though none of the methods use task independent lightweight encoder model as used in this work. Pohlen et al. \cite{pohlen2017fullresolution} proposed a ResNet like architecture that considers multi-scale input in two stream process. One stream extracts features from full resolution to get knowledge about semantic boundaries and the other one works on different dimension as received from sequence of pooling layers to extract robust local features. ENet \cite{paszke2016enet} presented a network aimed for mobile devices where ResNet like architecture is considered as encoder and it contains dilated convolution  to extract semantic features. Another realtime segmentation work, ICNet \cite{zhao2017icnet} reported a multi-branch architecture where inputs in different resolution are considered and it developed cascade feature fusion unit across multiple branches. In one more recent paper, ContextNet \cite{poudel2018contextnet} proposed multi-column encoder architecture and later fusion at decoder. A deep network as encoder is used to extract contextual features from relatively smaller resolution and another encoder that is shallow in nature is added to extract global features.

The methods mentioned above do not use task independent lightweight encoder model as used in this work and do not support multi task learning practice.

\section{Proposed Method} \label{proposed}

\subsection{Overview}

The reported work is conceptualized for the embedded platform where the entire system can run on real time, thus priority is execution speed over accuracy. Similar to existing neural network architectures such as FCN \cite{long2015fully}, SegNet \cite{badrinarayanan2015segnet}, UNet \cite{ronneberger2015u} for segmentation, the proposed network has also two parts which are encoder and decoder, which is furnished in Figure 1. An encoder is basically a Convolutional Neural Network (CNN) that helps to extract features from various feature dimensions and thus reduces the problem space as it becomes deeper. A decoder does the opposite of encoder, it consumes the feature maps from last layer as well as intermediate layers of encoder and reconstructs the original input space. For reconstruction we use deconvolution layer instead max-unpooling as used in SegNet \cite{badrinarayanan2015segnet} and ENet \cite{paszke2016enet}. Once the reconstruction is done then the decoder generates class prediction for each pixel in original input space.

\subsection{Encoder}

It is mentioned earlier that the intention of this study is to develop a lightweight decoder for segmentation task where the encoder is task independent. It means that the features extracted by the single encoder will be shared across all decoders to accomplish separate tasks (segmentation, detection, classification etc.). In the context of our work, it is observed that the encoder design is mostly specific to the semantic segmentation task such as \cite{badrinarayanan2015segnet}, \cite{romera2017efficient}, \cite{treml2016speeding}, \cite{yu2015multi}, thus these encoders had all necessary components to extract semantic features and the decoder performs well with the readily available information. However, our study is intended towards use of a more generic encoder. So we designed a task independent, designed a VGG \cite{simonyan2014very} style classifier of 10 layers as encoder and no special attention given for semantic segmentation such as heavy usage of dilated convolution as used in \cite{chen2018deeplab} \cite{chen2018encoder} \cite{DBLP:journals/corr/abs-1709-00179} \cite{romera2017efficient} \cite{wang2018understanding} \cite{yu2015multi} \cite{zhao2017icnet} \cite{yang2018denseaspp} or information fusion at different resolution \cite{lin2017refinenet} \cite{poudel2018contextnet} \cite{zhao2017icnet} \cite{zhang2018exfuse}.

As per Figure \ref{fig:EncoderDecoderDilated}, convolution with stride 2 followed by max-pooling is used to reduce the problem space, thus reducing the number of hyperparameters and also the run-time. Obviously, this is a trade-off for segmentation accuracy, but that is not the case for other tasks like detection, classification etc. Considering this encoder to be function independent, this gap of spatial information exploration needs to be overcome in the decoder side by learning semantic features extensively. Convolution layers are added sequentially along with increasing width and decreasing feature space in regular interval. All the convolution layers use kernel of size 5X5 followed by Batch-Normalization \cite{ioffe2015batch} to speed up the convergence and ReLU \cite{nair2010rectified} to add non-linearity. The last layer of encoder produces total 256 feature maps.

\subsection{Decoder}

The proposed architecture of decoder has been obtained after redesigning the existing features of the convolutional nets, such as residual learning \cite{he2016deep} and non-bottleneck layers \cite{romera2017efficient}. In the recent past, learning through residual blocks has shown breakthrough performance in many vision related tasks and made it possible to make the network grow more deeper very easily. This advancement helped to outperform significantly in many object recognition tasks. Further the same learning strategy has been also used for semantic segmentation as in \cite{romera2017efficient}. However, in \cite{dasevaluation}, it has been demonstrated that residual learning for a network with lesser depth is not efficient. This is because the network can not handle high gradient flow during back-propagation. To circumvent this issue, in this study, the original residual learning \cite{he2016deep} strategy has been modified with an adequate arrangement to distribute the high gradients through multiple skip connections.

As per our design of the encoder, it can be realized that there has been no suitable mechanism employed to extract semantic features. Now, following with the recent trend, if the decoder is going to have only few set of deconvolution layers to reconstruct the output then the segmentation result will be definitely poor. The reason is the performance of the decoder is limited to the knowledge shared by the encoder and as per our design, the encoder is expected to share knowledge that is non-semantic in nature. To address this affair, there is a requisite to learn semantic information from the encoded feature space in the decoder. Hence, we need non-bottleneck blocks between two deconvolution layers. In \cite{romera2017efficient}, the authors have designed non-bottleneck layer which is 1D in nature and claimed to be a better regularizer and also faster. The same non-bottleneck layer is used in the encoder as well and most of the convolutions are dilated.

\begin{figure*}[h]
    \centering
    \includegraphics[scale=0.38]{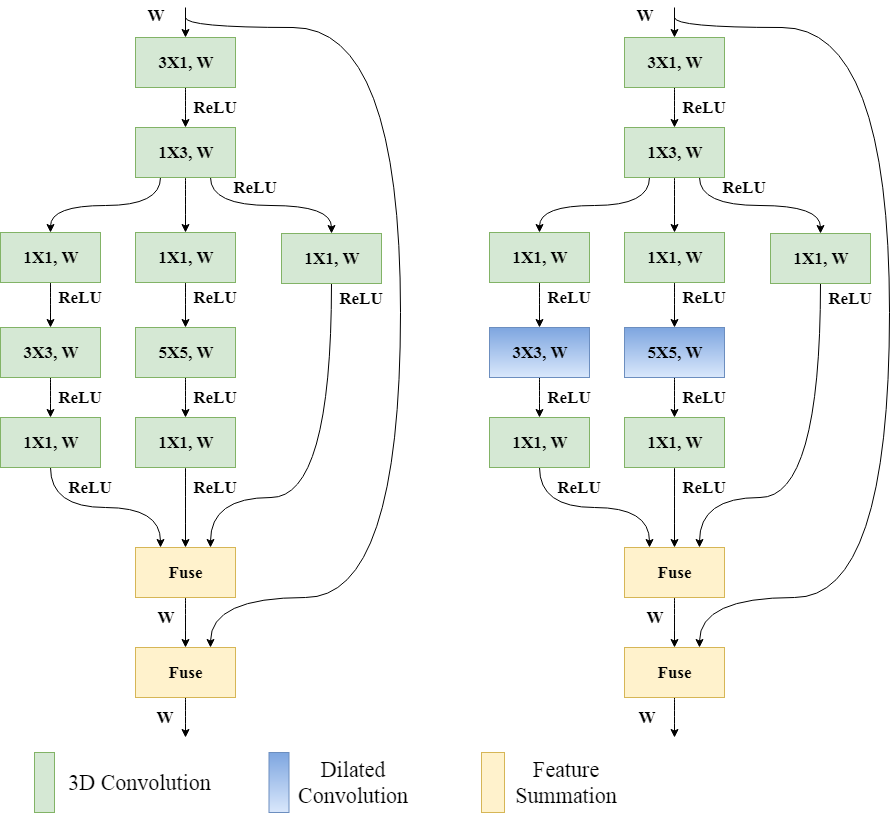}
    \caption{Illustration of non-bottleneck layers}
    \label{fig:Non-bottleneckAll}
\end{figure*}

The design of our non-bottleneck layer is shown in Figure \ref{fig:Non-bottleneckAll}. It contains both 1D and 3D kernels. 1D kernel is used to extract information mainly in one direction at a time and 3D kernel is for gathering features from bigger receptive area. Later we try to look for dense information through multiple kernels with different sizes for example 3X3, 5X5 and 1X1 respectively. Following this, the features extracted using different kernels are fused. This method helps to summarize the semantic features that are collected from different receptive area. The resultant features are again fused with the input features to the same non-bottleneck layer. The multiple skip connections to the feature fusion blocks in the proposed non-bottleneck layer help to handle high gradient flow because during back-propagation the incoming gradient gets distributed among all paths. As per Figure \ref{fig:Non-bottleneckAll}, our non-bottleneck layer has two variants that are type-1 (left) and type-2 (right). The only difference between two variants is the block at the right uses dilated convolution where the kernel size is 3X3 and 5X5. Dilated convolution helps to extract spatial information by expanding the field-of-view as per the dilation factor while maintaining the same resolution. With increasing dilation rate, receptive field is also expanded however, for the present work the dilation rate has been kept constant and it is 1. It is possible to receive better accuracy with incremental dilation rate but dilated convolution is computationally expensive and it becomes more costlier as the dilation rate increases. Considering the present study to be aimed for embedded platform, usage of similar features need to be employed carefully to meet the run-time. After each convolution operation, ReLU \cite{nair2010rectified} is used as activation unit for better convergence \cite{krizhevsky2012imagenet}.

\subsection{Encoder-Decoder architecture}

\begin{figure*}
\centering
\includegraphics[scale= 0.45]{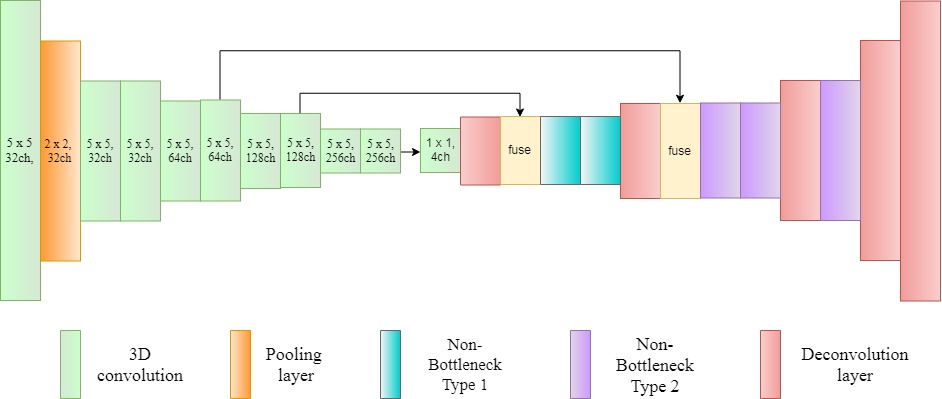}
\caption{CNN based encoder-decoder architecture used for semantic segmentation task}
\label{fig:EncoderDecoderDilated}
\end{figure*}

Figure \ref{fig:EncoderDecoderDilated} shows the encoder-decoder pair that is used in this work. The encoder is a very generic one and the decoder is our main proposal. The feature maps passed from the encoder are downsampled from 256 to 4 as the present study concentrates on 4 classes. One can argue that drastic change in the number of feature maps such as this can impact on accuracy but if we make the decoder wider that will shoot up the runtime significantly. Thus decreasing the number of feature maps in regular interval is not affordable and out of our budget. After first deconvolution layer, non-bottleneck layer of type-1 is used two times. Following second and third deconvolution layers, non-bottleneck layers of type-2 are used twice and once respectively. There is no non-bottleneck layer used after fourth deconvolution. The non-bottleneck layers are designed in such a way that first repetitive usage of type-1 as intermediate block between two deconvolution layers and then having multiple type-2 especially between the later deconvolution layers will help to remove the gap of learning semantic information that we have seen in the encoder side. Hence, our proposed non-bottleneck layer is generic and can be used for any segmentation task. Also we pulled intermediate feature maps from the second and third last convolution layers of the encoder. This has to be highlighted that our encoder consumed most of the run time because it is much wider than the decoder. To circumvent this issue, we could use the concept of group convolution as proposed by Xie et \textit{al.} \cite{xie2017aggregated} but in \cite{dasevaluation2018}, it has been experimentally demonstrated that aggregated residual transformation shows adverse effect for lightweight networks such as our encoder.

Details on how to train our proposed network end-to-end, other training strategy and hyperparameter details, are discussed in the next section.

\begin{figure*}[ht!]
\centering
\begin{subfigure}{.48\textwidth}
    \includegraphics[width=\textwidth]{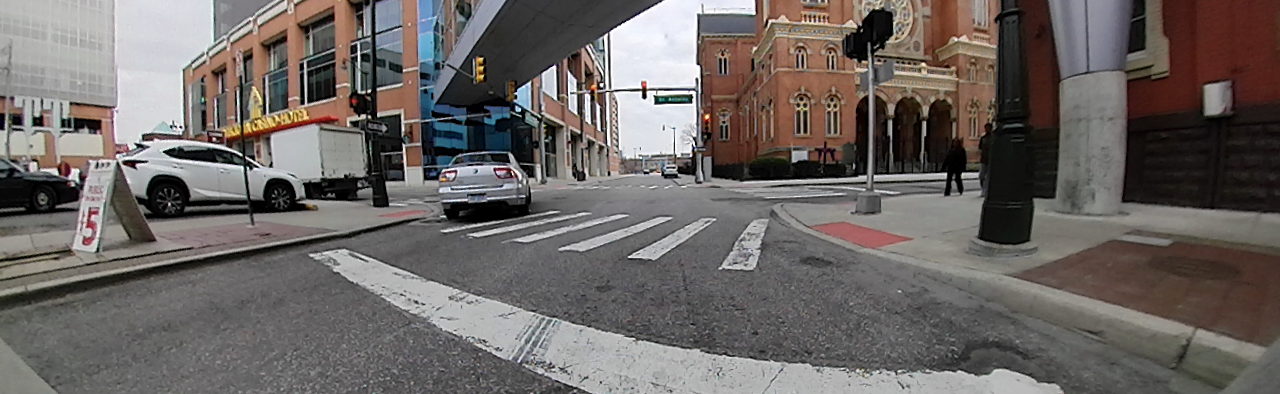}
    \caption{\textcolor{black}{Input Image - Example 1} \newline}
\end{subfigure}%
\quad
\begin{subfigure}{.48\textwidth}
    \includegraphics[width=\textwidth]{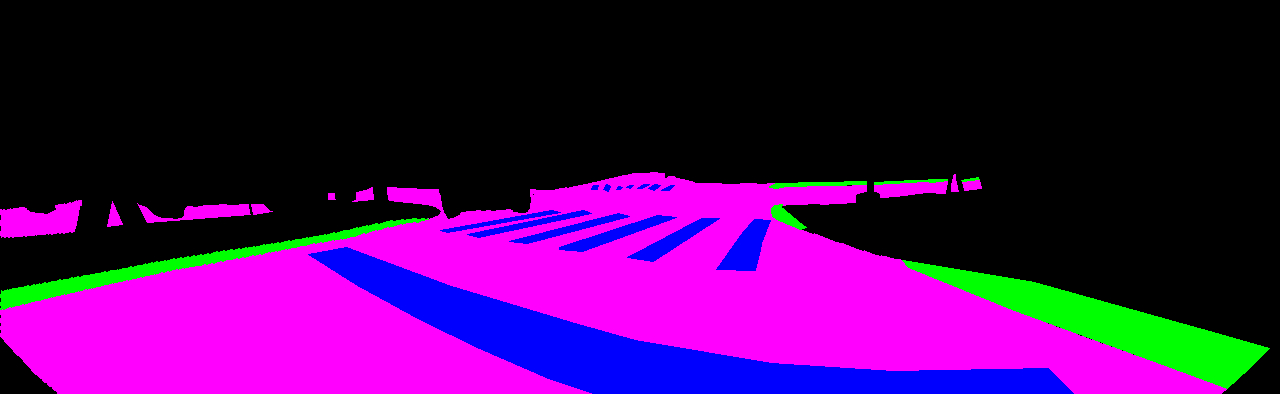}
    \caption{\textcolor{black}{Ground Truth - Example 1}\newline}
\end{subfigure}%
\quad
\begin{subfigure}{.48\textwidth}
    \includegraphics[width=\textwidth]{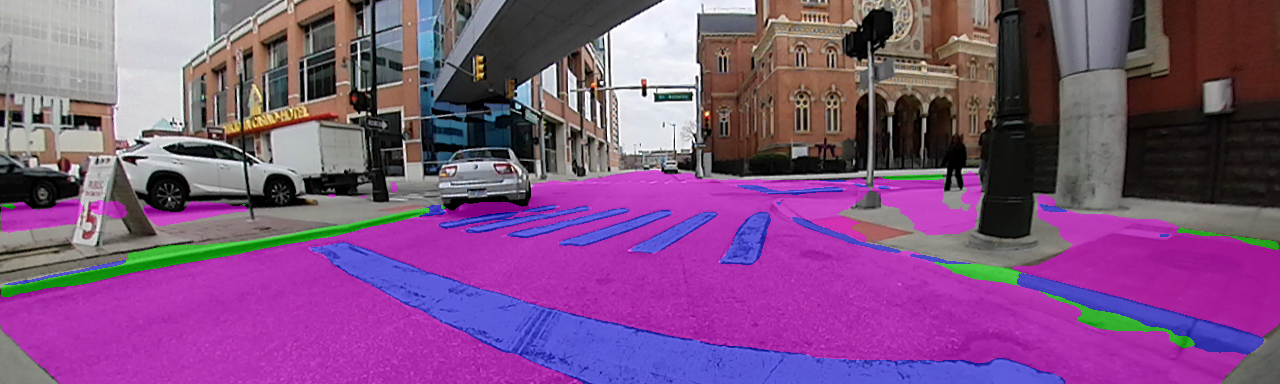}
    \caption{\textcolor{black}{FCN Output - Example 1}\newline}
\end{subfigure}%
\quad
\begin{subfigure}{.48\textwidth}
    \includegraphics[width=\textwidth]{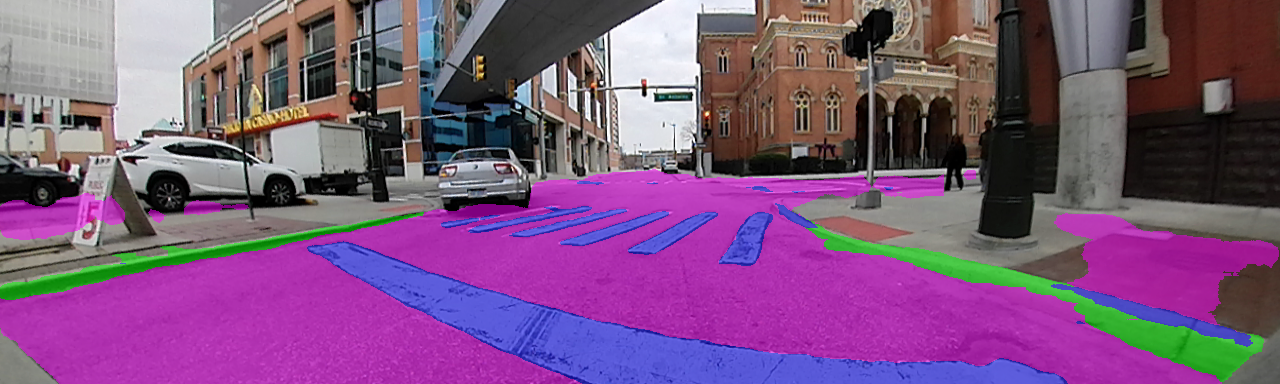}
    \caption{\textcolor{black}{Proposed Decoder Output - Example 1}\newline}
\end{subfigure}
\quad
\begin{subfigure}{.48\textwidth}
    \includegraphics[width=\textwidth]{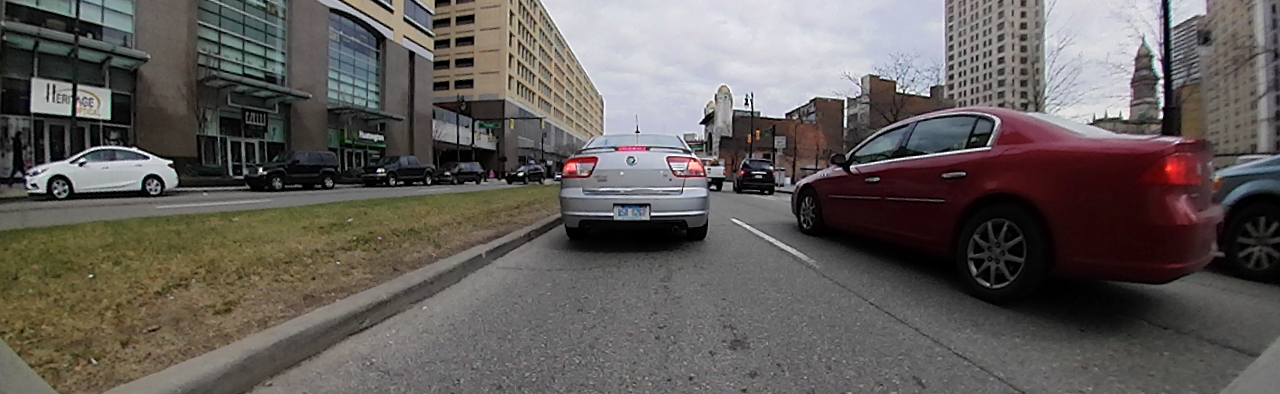}
    \caption{\textcolor{black}{Input Image - Example 2} \newline}
\end{subfigure}%
\quad
\begin{subfigure}{.48\textwidth}
    \includegraphics[width=\textwidth]{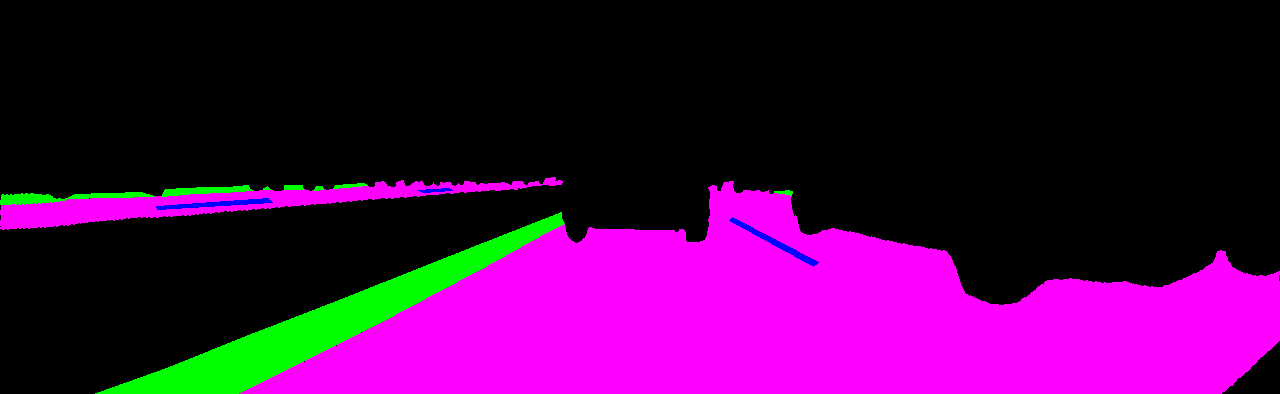}
    \caption{\textcolor{black}{Ground Truth - Example 2}\newline}
\end{subfigure}%
\quad
\begin{subfigure}{.48\textwidth}
    \includegraphics[width=\textwidth]{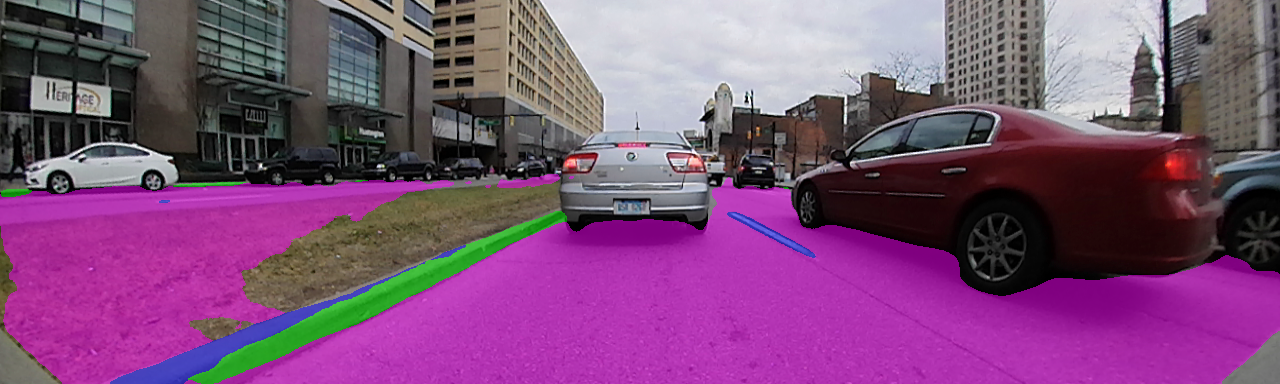}
    \caption{\textcolor{black}{FCN Output - Example 2}\newline}
\end{subfigure}%
\quad
\begin{subfigure}{.48\textwidth}
    \includegraphics[width=\textwidth]{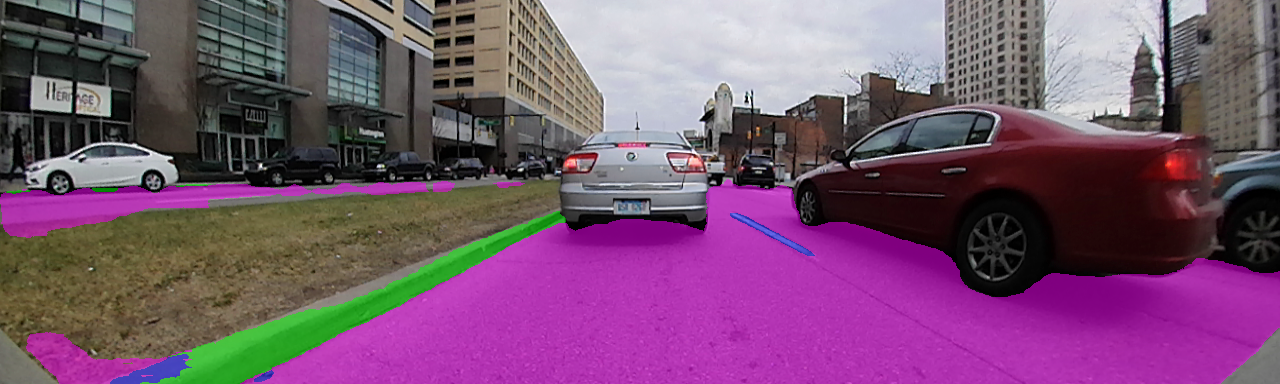}
    \caption{\textcolor{black}{Proposed Decoder Output - Example 2}\newline}
\end{subfigure}

\caption{Comparison of proposed optimal decoder against a standard FCN decoder}
    \label{fig:ResultComparisonX}
\end{figure*}

\section{Experiments} \label{results}

We have performed thorough investigation to show the efficacy and robustness of our proposed network.

\paragraph{Dataset}

We have used dataset that is owned by our organization. However, some samples of the dataset are shown in Figure 4. In this dataset, dimension of each image is 1280X384. Samples are highly varied in nature and mostly urban scenes. Diverse lighting conditions and presence of dense shadow make this dataset challenging. In the present study, to perform semantic segmentation, we concentrated only on 4 critical classes that are lanes, curb, road and void (everything else). The entire dataset is divided into Training, Validation and Test set each containing 3016, 981 and 1002 images respectively. Corresponding each sample image, single channel annotation was developed where each pixel has only one class label.

\begin{table*}
  \caption{Results of different variants of our decoder with non-bottleneck layer and VGG10 as encoder for semantic segmentation on our dataset }
  \label{table1}
  \centering
  %\scriptsize
  \begin{tabular}{llllllll}
    \toprule
    \multicolumn{6}{c}{}{Avg. class accuracy} {Avg. class IoU score}\\
    \cmidrule(r){2-4}
    \cmidrule(r){5-7}
    Decoder configuration   &Lanes &Curb &Road &Lanes &Curb &Road &Mean \\
    \midrule
    D1     &0.5678  &0.6228  &0.9688 &0.4943 &0.4296 &0.9265 &0.7255 \\
    D2     &0.4363  &0.6082  &\textbf{0.9694} &0.4021 &0.4445 &0.9198 &0.6989 \\
    D3     &0.5096  &0.5682  &0.9689 &0.4333 &0.4264 &0.9254 &0.7038 \\
    D4     &0.4769  &0.5932  &0.9684 &0.4152 &0.4534 &0.9248 &0.7013 \\
    D5     &0.4561  &0.6059  &0.9668 &0.4013 &0.403 &0.9138 &0.6866  \\
    D6     &0.5775  &0.6539  &0.9566 &0.4567 &0.4198 &0.9154 &0.7106 \\
    D7     &0.5263  &0.4284  &0.9464 &0.4192 &0.2911 &0.8689 &0.6458 \\    
    D8     &0.5628  &\textbf{0.6755}  &0.9686 &0.4752 &0.4717 &0.9296 &0.727 \\
    (2N1)(2N1)(2N2)(2N2)    &0.6054  & 0.6534 &0.9685 &0.5294 &\textbf{0.4755} &0.9299 &0.742 \\
    (2N2)(2N2)(2N2)(2N2)    &0.5951  &0.647   &0.9655 &0.5061 &0.5061 &0.9267 &0.7323 \\
    (2N2)(2N2)(2N2)         &0.6     &0.639   &0.9661 &0.5063 &0.4391 &0.9279 &0.7315 \\
    (2N1)(2N1)(2N1)(2N1)    &0.5743  &0.6343  &0.9691 &0.4961 &0.4486 &0.9275 &0.7284 \\
    (1N1-1N2)(1N1-1N2)(1N1-1N2)(1N1-1N2) &0.5938 &0.6338 &0.9681 &0.5051 &0.4438 &0.9264 &0.731\\
    Optimal  &\textbf{0.6118} &0.6588 &0.9689 &\textbf{0.5304} &0.4696 &\textbf{0.9314} &\textbf{0.7441}\\
    \bottomrule
  \end{tabular}
\end{table*}

\paragraph{Training}

First we train our encoder from scratch on ImageNet and then transfer the weights for the present task. This transfer learning will help to get better initialization of the weights at the beginning of the training for semantic segmentation. The pre-training on a much larger dataset was required because our model is quite shallow and the dataset used in this work is very small in size. Consequently there is a high chance that the network will suffer from over-fitting due to lack of training samples. We could follow the concept of layer-wise training in a supervised fashion to train the encoder to extract more robust features as reported in \cite{roy2016generalized} as the encoder used in this work is quite lightweight. Implementation of the proposed network and all experiments are executed using Keras \cite{chollet2015keras} framework. We considered very popular Adam \cite{kingma2014adam} as optimizer.  Regarding the other network configuration, weight decay and batch size were set to 0.9 and 4 respectively. Training was started with 0.0005 as initial learning rate including standard polynomial decay strategy to decrease this value over 350K iterations. Dropout \cite{srivastava2014dropout} is not used in our model. For all experiments, we used NVIDIA Titan X 12G GPU with 24 GB RAM. No data augmentation technique has been performed during training.

\paragraph{Experimental results and comparison study}

The hardware that we use is designed with automotive power constraints in mind, thus having restricted number of features for design of a convolutional neural network. Also we intend to utilize a generic encoder network that stays well within the budget.Thus our main objective is to design an efficient decoder that satisfies both these constraints. In the course of design of an efficient decoder, we have experimented multiple versions of decoder all of which are explained later in this section. With all these decoders VGG10 pre-trained with Imagenet is used as the encoder. This will help to have fair comparison of the different variants of the decoder. The entire network containing the encoder along with different decoders is trained end to end with the available pixelwise ground truth label. All these variants of decoder reported in Table \ref{table1}, fits all our constraints.   %Among several existing segmentation networks, we have considered FCN   to compare the efficacy and robustness of our proposal. VGG10 as described in the earlier section has been considered as the encoder and FCN (with kernel size of 5X5) is paired as decoder to complete the network. This will help to have a fair comparison among FCN and the proposed decoder because the encoder is the same. So basically we have two networks (VGG10+FCN) and (VGG10+proposed decoder) which are trained with the same strategy of pre-training encoder on ImageNet, followed by entire network training and obtaining pixel level accuracy as shown in Table 1. We estimated run time for both networks on our embedded platform and our proposed network takes almost 2 millisecond extra time than VGG10+FCN network. The reason is  that the proposed decoder contains non-bottleneck layers that use a good number of 1X1 convolutions and this operation is computationally expensive. However, this cost  can be justified especially for the classes that are not well represented in the dataset.% 
It is quite general while capturing an urban scene, there will be very limited region occupied by lanes, curbs but it is exactly opposite for roads and void classes. So we clearly see that for effective learning there is a huge gap in problem space in these two classes while comparing with other. Without even attempting any data augmentation and class weighing technique, our non-bottleneck layers worked better for curb though there is a slight deterioration for lanes. To evaluate the segmentation performance on all the designed decoders, widely used Intersection over Union (IoU) metric is considered and details are furnished in Table \ref{table1}. 

As put forth earlier, we did experiments with several combinations of Non-Bottleneck layers in the network whose results are updated in Table \ref{table1}. Decoder D1 uses our proposed non-bottleneck layer without 1X1 convolution after 3X3 and 5X5 convolution. Decoder D2 is same as D1 but it does not use second skip connection from encoder. Decoder D3 shares the same configuration as D2 but the batch size during training was 8 whereas it was 4 for D2. Decoder D4 is same as D3 but it does not use 1X1 convolution even before 3X3 and 5X5 convolution. Decoder D5 is a bit different. After first deconvolution layer two sets of 3X1, 1X3 convolutions followed by ReLU is used. Also it uses skip connection to fuse the resultant features with the input feature maps of first 3X1 convolution. After second deconvolution layer, one 3X3 dilated convolution with dilation rate 1 is used and then the same non-bottleneck as used after first deconvolution layer. Only 3X3 dilated convolution with dilation rate 1 is used after third and fourth deconvolution layer. Decoder D6 is same as D5 but it uses batch size as 4 where 8 was used in D5. Decoder D7 is different in terms of kernel size in deconvolution layers. It uses kernel of size 2X2 in first and second, 3X3 in third and fourth, 5X5 in fifth upsampling layer. Decoder D8 uses same non-bottleneck as D7 without 3X3 and 5X5 convolution. In the pattern \textit{m}N\textit{p} in Table \ref{table1}, \textit{m}  stands for number of non-Bottleneck (N) layers, \textit{p} stands for the type of non-bottleneck layer. Representation within braces ( and ) stands for set of non-bottleneck layers after a  deconvolution layer starting from the first one. Further to explore more, we have modified the design of our non-bottleneck layer from various aspects to check what variant of change seems to work better. Please note that the modified non-bottleneck layers are non-repetitive between two deconvolution layers. Of the different decoder variant, the best version is the one put forth in Figure \ref{fig:EncoderDecoderDilated}, which is obtained after several optimization efforts and this network uses the Non-bottleneck layers detailed in Figure \ref{fig:Non-bottleneckAll}. This network also takes care of the class imbalance for Lanes and Curb and improves its class-wise IoU. We also compared the runtime measurement of this network with the popular FCN network on our custom hardware. We find that this network has the advantage of improving the IoUs especially of the key classes with less than 2ms increase in runtime. The sample segmentation outputs of our proposed optimal decoder and a standard FCN decoder are shown in Figure \ref{fig:ResultComparisonX}.

%\begin{table*}
% \caption{Performance comparison between FCN \cite{long2015fully} and proposed method }
%  \label{table3}
%  \centering
%  \begin{tabular}{lllllllllll}
%    \toprule
%    Network      &Avg. test accuracy &Avg. precision &Avg. recall &Avg. F1 score\\
%    \midrule
%    D1           &0.9681             &0.7615         &0.8256      &0.7785 \\
%    D2           &0.965              &0.7284         &0.8401      &0.7579 \\
%    D3           &0.9682             &0.7407         &0.8243      &0.7606 \\
%    D4           &0.9675             &0.7421         &0.828       &0.7615 \\
%    D5           &0.9626             &0.7329         &0.7985      &0.7459 \\
%    D6           &0.9634             &0.7671         &0.7861      &0.7654 \\
%    D7           &0.9431             &0.6935         &0.7866      &0.71 \\
%    D8           &0.9694             &0.7786         &0.8144      &0.7843 \\
%    (2N1)(2N1)(2N2)(2N2)   &0.9697      &0.7829     &0.841      &0.7983 \\
%    (2N2)(2N2)(2N2)(2N2)   &0.9683      &0.7736     &0.8219      &0.7858 \\
%    (2N2)(2N2)(2N2)        &0.9687      &0.773      &0.8208      &0.7839 \\
%    (2N1)(2N1)(2N1)(2N1)   &0.9686      &0.7685     &0.831       &0.7839 \\
%    (1N1-1N2)(1N1-1N2)(1N1-1N2)(1N1-1N2) &0.9682      &0.7712     &0.8249       &0.7842 \\
%    VGG10+Proposed         &0.9703      &0.7841     &0.8375       &0.7975 \\
%    \bottomrule
%  \end{tabular}
%\end{table*}

\section{Conclusion} \label{conclusion}
Design of efficient encoders is a growing area of research. In this work, we focused on design of efficient decoders. Firstly, we designed a novel efficient non-bottleneck layer and a family of decoders based on this layer. We experimentally show that different choice of decoder design had a large impact and the optimal configuration had 10\% improvement of accuracy in terms of mean IoU over a baseline configuration. In particular, for more difficult segmentation classes like lanes and curb, higher improvements of 12\% and 18\% were obtained. Thus we demonstrate that the design of an efficient decoder can play a critical role for segmentation tasks as it covers a significant portion of the overall computation of the network. We hope that our preliminary study demonstrates the need for further research on efficient decoders. In future work, we build a systematic family of meta-architectures with a fixed run-time budget and learn the optimal configuration using meta-learning techniques.


% acknowledgments go at the end of the paper. Do not include
% acknowledgments in the anonymized submission, only in the final paper.

% {
% \medskip
% \small
% \bibliographystyle{plain}
% \bibliography{references}
% }

%\vfill
\bibliographystyle{apalike}
{\small
\bibliography{references}
}

% \section*{\uppercase{Appendix}}

% \noindent If any, the appendix should appear directly after the
% references without numbering, and not on a new page. To do so please use the following command:
% \textit{$\backslash$section*\{APPENDIX\}}

\vfill
\end{document}